# TIBECG: Temporal Information Bottleneck for Lead-Specific Demographic Privacy in ECG Waveform Release

Tsai-Min Chen[1,2] and Jun Seita[2]

[1] Research Center for Information Technology Innovation (CITI), Academia Sinica, Taipei 115201, Taiwan

[2] RIKEN Center for Interdisciplinary Theoretical and Mathematical Sciences (iTHEMS), RIKEN, Tokyo 103-0027, Japan

Corresponding author: Tsai-Min Chen, RIKEN Visiting Scientist and Academia Sinica Postdoctoral Scholar. ORCID 0000-0002-1143-0677

Abstract—Transmitting an electrocardiogram (ECG) waveform can expose demographic information beyond an intended diagnostic task. We evaluate TIBECG, an architectural temporal information bottleneck, on individual leads extracted from hospital twelve-lead recordings. A 2,302,337-parameter processor varies its latent length while retaining the same nine-score consistency objective, without age or sex labels in protector fitting. Across 6,386 recordings, ten folds, 12 leads and 21 lengths, retrospective selections retained at least 95% of each lead's clean macro-F1. Their weaker age/sex interval gain exceeded the best eligible conventional transformation on all leads, by a median 22.08 percentage points. The implemented sampling-rate comparison shows that direct downsampling usually suppresses demographic inference more strongly but at a larger loss of diagnostic utility, whereas TIBECG favors utility preservation. Selection stability also varies by lead. Lower latent Kullback-Leibler (KL) divergence does not reliably rank residual sex leakage at a fixed length. These findings support lead-specific evaluation of waveform processing, not validated consumer deployment or expert-readable reconstruction. When the waveform-input diagnostic classifier is hosted as part of a proprietary cloud service rather than deployed on the device, waveform transmission is required to obtain its scores.

## I. Introduction

Consumer electrocardiogram (ECG) systems can release diagnostic scores or waveforms. Wireless biopotential acquisition [1] and residential ECG networks [2] support remote ECG processing. A related homecare platform processes phonocardiograms (heart sounds) [3]. In each setting, the released object determines what the recipient can infer. Scores suffice when the intended classifier can run locally. In the deployment setting considered here, however, the classifier remains a proprietary cloud-service asset and accepts the waveform remotely; the device must therefore release a waveform to obtain the diagnostic result.

Tangari et al. found code capable of collecting personal data in 88.0% of the mobile health apps examined [4, p. 1]. ECG-based age and sex estimation demonstrates a separate exposure route even when explicit demographics are withheld [5]. We therefore study demographic inference conditional on waveform release, without assuming that every consumer service uses this deployment model.

Building on SRECG and its high-resolution multiclass classifier (HMC) [6, pp. 252-253, Fig. 2], TIBECG changes internal time resolution without changing parameter count or the utility-only objective. We evaluate individual clinical leads, compare architectural restriction with conventional transformations and sampled inputs, and test selection sensitivity. This is a retrospective engineering study; consumer acquisition and clinical interpretation remain validation targets.

## II. Related work

### *A. The signal a consumer service needs*

Consumer systems support point-of-care ECG classification [7], remote monitoring [8] and human-tested arrhythmia detection [9]. In Raj's prototype, predicted heartbeat classes are transmitted to a remote receiver [8, p. 111]. That design performs inference before transmission. We study the complementary deployment setting in which a proprietary classifier remains at the remote service; one frozen HMC maps a transmitted ten-second single-lead waveform to nine scores. The experimental utility endpoint is its macro-F1, not the full set of possible clinical uses of a trace. Critical-beat detection [10] and graph-based heartbeat classification [11] target specific events.

ECG compression [12] and wearable artifact reduction [13] aim to preserve useful signal features. Those features can also encode demographics. We therefore evaluate the same released object with the frozen diagnostic classifier and separately trained age and sex auditors.

### *B. The point at which privacy is evaluated*

Edge seizure detection [14], wearable EEG monitoring [15], local stress estimation [16] and sleep-derived stress sharing [17] illustrate task-specific output choices. Authentication [18], signatures [19] and access control [20] protect communication or permissions. Our threat model begins after an authorized cloud service hosting the proprietary classifier has obtained the released waveform.

Federated learning [21] and personalized wearable representations [22] alter the training or collaboration setting. HealthGuard uses temporal randomization and server-side reconstruction [23, p. 1465]. ECG demographic adversaries [24] and adversarial representation censoring [25] directly model sensitive targets. TIBECG instead fits the protector using only diagnostic scores; any new sensitive target requires a new audit, and protection of that target is not guaranteed.

## III. Data and waveform processing

### *A. Cohort and diagnostic teacher*

The CPSC2018 public training set contains 6,877 clinical twelve-lead ECGs sampled at 500 Hz [26]. We required one diagnostic label, finite waveforms, known age and recorded binary sex, and at least 4,999 samples. We used the terminal nominal ten-second segment, allowing a one-sample shortfall at 500 Hz. Polyphase resampling yielded 2,500 points at 250 Hz without explicitly padding the input array; filtering used the implementation's default zero extension at the boundaries. The 6,386 eligible records comprised normal rhythm (915), atrial fibrillation (974), first-degree atrioventricular block (686), left and right bundle branch block (178 and 1,530), premature atrial and ventricular contractions (530 and 606), and ST depression and elevation (782 and 185).

Ten diagnosis-stratified folds were shared across leads and conditions. The test fold, preceding cyclic validation fold and remaining eight training folds were disjoint at recording level. Each record was tested once; patient independence could not be established from the available identifiers. Each HMC, protector and auditor sees one lead, not all 12 simultaneously. This channel-wise protocol does not reproduce consumer electrodes, motion or device filtering. The analysis reuses public CPSC2018 data; no new recordings were acquired.

The HMC uses five convolutional blocks, each with two width-three convolutions and a strided convolution, followed by a bidirectional gated recurrent unit (GRU), attention and a nine-score sigmoid head. This design follows our arrhythmia classifier [27, Fig. 1]. Twelve channels per convolutional block and 12 recurrent units per direction give 27,609 trainable parameters in this PyTorch implementation; SRECG reports 28,035 in its Keras implementation [6, p. 252]. Each lead and fold has its own HMC. We used diagnosis labels, categorical cross-entropy on normalized sigmoid scores, Adam at 0.001, batch size 32 and validation-loss checkpoint selection. The teacher weights were frozen for protection training.

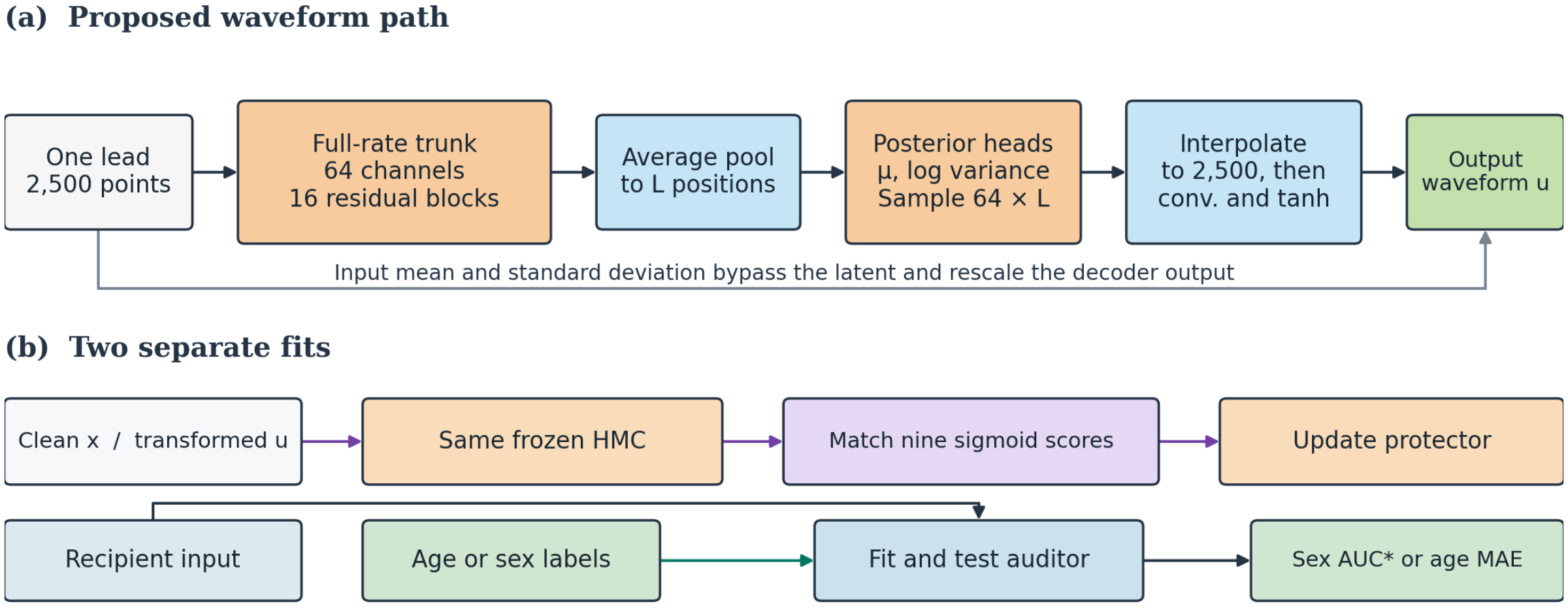

**Fig. 1.** Proposed waveform-processing path and experimental supervision. In (a), orange marks the trunk and posterior heads; blue marks temporal pooling and output processing, which includes a trainable convolution. In (b), the HMC represents the frozen cloud-service classifier while the protector is updated. The input mean and scale bypass the latent. Demographic labels determine cohort eligibility and supervise the separate audit, but do not supervise protector fitting. Device placement is proposed, not evaluated on hardware.

### *B. Restricting internal time resolution*

Fig. 1 shows where temporal resolution changes. The encoder applies a width-16 convolution, 16 residual blocks with two convolutions and batch normalization each, and a long skip connection. Adaptive average pooling maps 64 full-rate channels to L positions. Two width-16 heads return a posterior mean μ and log variance v, clipped to [−8, 4] for v. We use a Gaussian draw $z = \mu + \exp(v/2) \odot \varepsilon$, with standard-normal ε, as in the reparameterization of Kingma and Welling [28, arXiv v11, p. 5, Eq. (10)].

When L is shorter than 2,500, linear interpolation restores 2,500 time positions, followed by leaky rectification, an output convolution and a hyperbolic tangent producing y. The released waveform is $u = c(x) + 2s(x)y$, where c(x) is the input mean and s(x) is its sample standard deviation, floored at $10^{-6}$. The recipient receives 2,500 waveform samples. The latent and its coordinates stay inside the processor. A key, record identifier and latent shape specify the test draw. The experiment evaluates one release per record under this protocol.

For one record, the loss is the average squared difference between the nine raw sigmoid scores $h_j(u)$ and $h_j(x)$ of the frozen HMC. We average over the training batch. Adam used a learning rate of 0.0001, effective batch size 32 and 100 epochs in full precision. The lowest validation consistency loss selected the checkpoint. Ground-truth diagnoses supervised the earlier teacher. Age and sex were used for cohort eligibility and subsequent audits. They supplied neither gradients nor a checkpoint criterion for the protector.

We trained separate protectors for L = 50, 100 through 250 in steps of 10, 500, 1,000, 1,250 and 2,500. Pooling and interpolation add zero trainable weights, so all 21 models have 2,302,337 parameters. L = 2,500 retains the stochastic processor at full length. Initialization depends on the condition. The sweep holds capacity as measured by parameter count and the training objective fixed, while varying the number of latent scalars, 64L.

*C. What the auditors observe*

The clean reference C gives the recipient an unmodified lead. The output-only reference O gives the nine clean HMC scores. Their class decisions are identical. The waveform auditor follows the HMC backbone with a scalar head (27,409 parameters). The score-only auditor applies batch normalization, leaky rectification, dropout and a scalar head to nine scores (28 parameters). Thus the gap between C and O reflects both the representation and the fitted attacker. These two empirical references define the scale in Fig. 2.

Waveform and score-only auditors were fitted separately for each lead, fold and target, using three initializations per condition. Sex used binary cross-entropy. Age used squared error on ages standardized using the training split and was converted back to years. We used Adam at 0.0001, batch size 32 and the lowest validation loss during 100 epochs. Within each test fold we retained the largest AUC* = max(AUC, 1 − AUC) for sex and the smallest mean absolute error (MAE) for age. Higher AUC* and lower MAE indicate greater leakage. The orientation and three-seed envelope use test outcomes. Clean auditors additionally retried weak validation fits (sex AUC* below 0.60 or age MAE above 95% of the training-mean predictor's validation MAE).

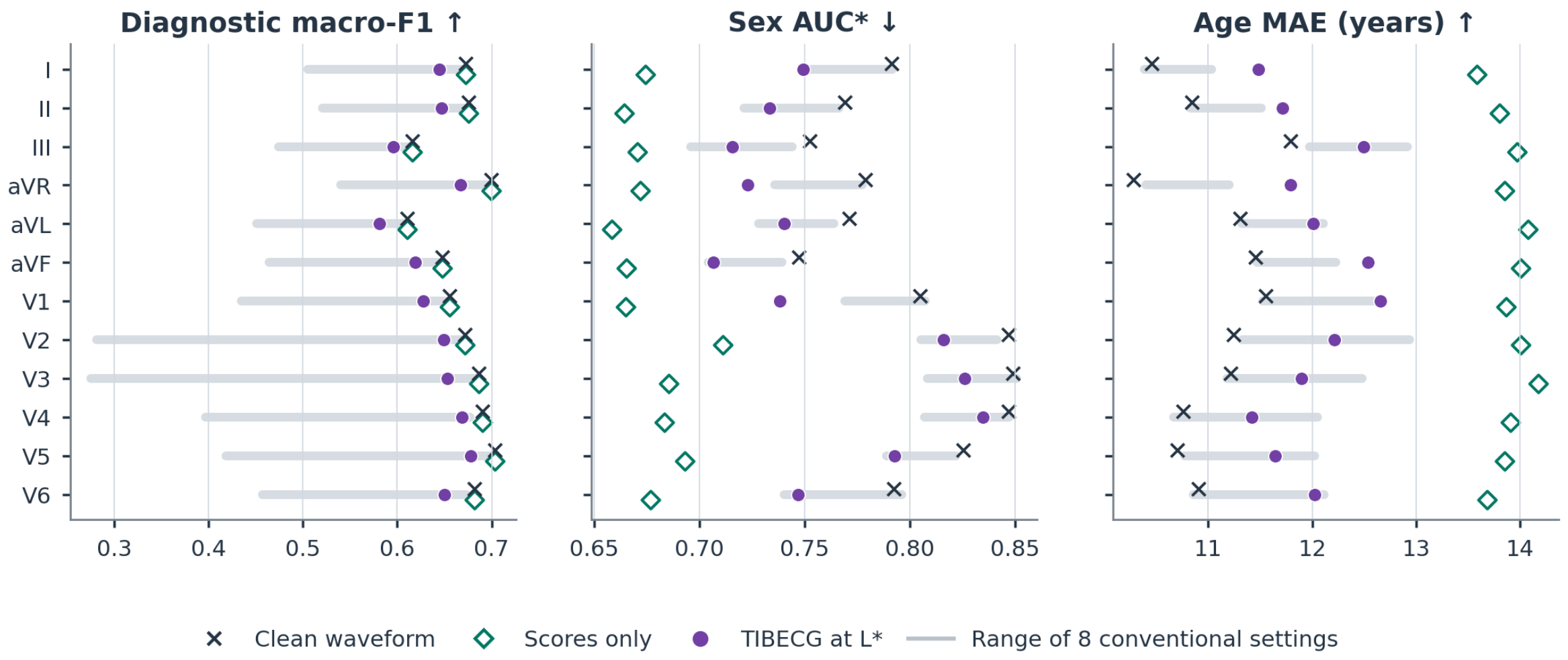

**Fig. 2.** Native-scale results by lead. Each point is a median of ten test-fold scores. The gray segment spans the eight conventional medians on that lead. Arrows indicate the preferred direction. Clean and output-only coincide in diagnostic macro-F1. Table S2 gives exact values for each method and each tested length.

## IV. Comparison and selection rules

*A. Eight conventional waveform transformations*

Each conventional transformation takes the same single-lead waveform and is evaluated with the same frozen HMC. A new auditor is fitted to each transformed condition. Gaussian noise has zero mean and standard deviation (SD) 0.25 or 0.50 times the record SD. Laplace noise has zero location and scale 0.15 or 0.30 times that SD (noise SD = √2 times scale). Independent draws are made per sample in a record-specific random stream. Three- and six-bit quantization map the record's amplitude range onto 8 and 64 levels, respectively, with a guard for a constant record. Block permutation shuffles contiguous one- or two-second blocks without reordering samples within each block.

*B. Percentages computed within each lead*

Let F be macro-F1, the unweighted mean over nine class F1 scores after argmax, S be sex AUC*, and E be age MAE in years. Each is first reduced to a median across ten folds for one lead and one condition. Subscripts C and O refer to the medians for the two reference inputs. Adapting the two-reference recovery measure from SRECG [6, p. 253, Eq. (8)], we use the following percentages:

$$U = 100\frac{F}{F_C}, \quad \Delta U = U - 100 \quad (1)$$

$$P_{sex} = 100\frac{S_C - S}{S_C - S_O} \quad (2)$$

$$P_{age} = 100\frac{E - E_C}{E_O - E_C} \quad (3)$$

U is the percentage of clean utility retained. The denominators require $F_C > 0$, $S_C > S_O$ and $E_O > E_C$, which hold for all 12 leads here. A clean-to-output utility interval would have a zero denominator. Each privacy gain is 0% at clean and 100% at output-only. Gains outside the interval are retained. A gain of 35% closes 35% of the observed gap between these attackers. It is neither an attack error rate nor a percentage of users protected. Cross-lead summaries take the median of 12 lead-level values with equal weight.

For example, the V4 sex auditor scored 0.84704 on clean, 0.68336 on scores only and 0.83469 at L = 500. The decrease was 0.01235 AUC against a reference gap of 0.16369, closing 7.55% of that gap using unrounded scores. On the same lead, the age error rose from 10.756 to 11.419 years against a score-only error of 13.904 years. This closes 21.07% of the age gap. Both percentages can therefore improve while the sex auditor remains discriminative.

*C. A joint, retrospective operating point*

For each lead, we retain measured lengths with U ≥ 95% and select the one with the highest J = min($P_{sex}$, $P_{age}$). Ties favor higher U, then shorter L. We denote the choice by L*. The same rule with one target yields the sex-only and age-only optima. The threshold is an engineering tolerance on macro-F1. The selection and its reported performance use the same test-fold summaries. We separately perturb the tolerance and omit folds to measure stability in Section VIII.

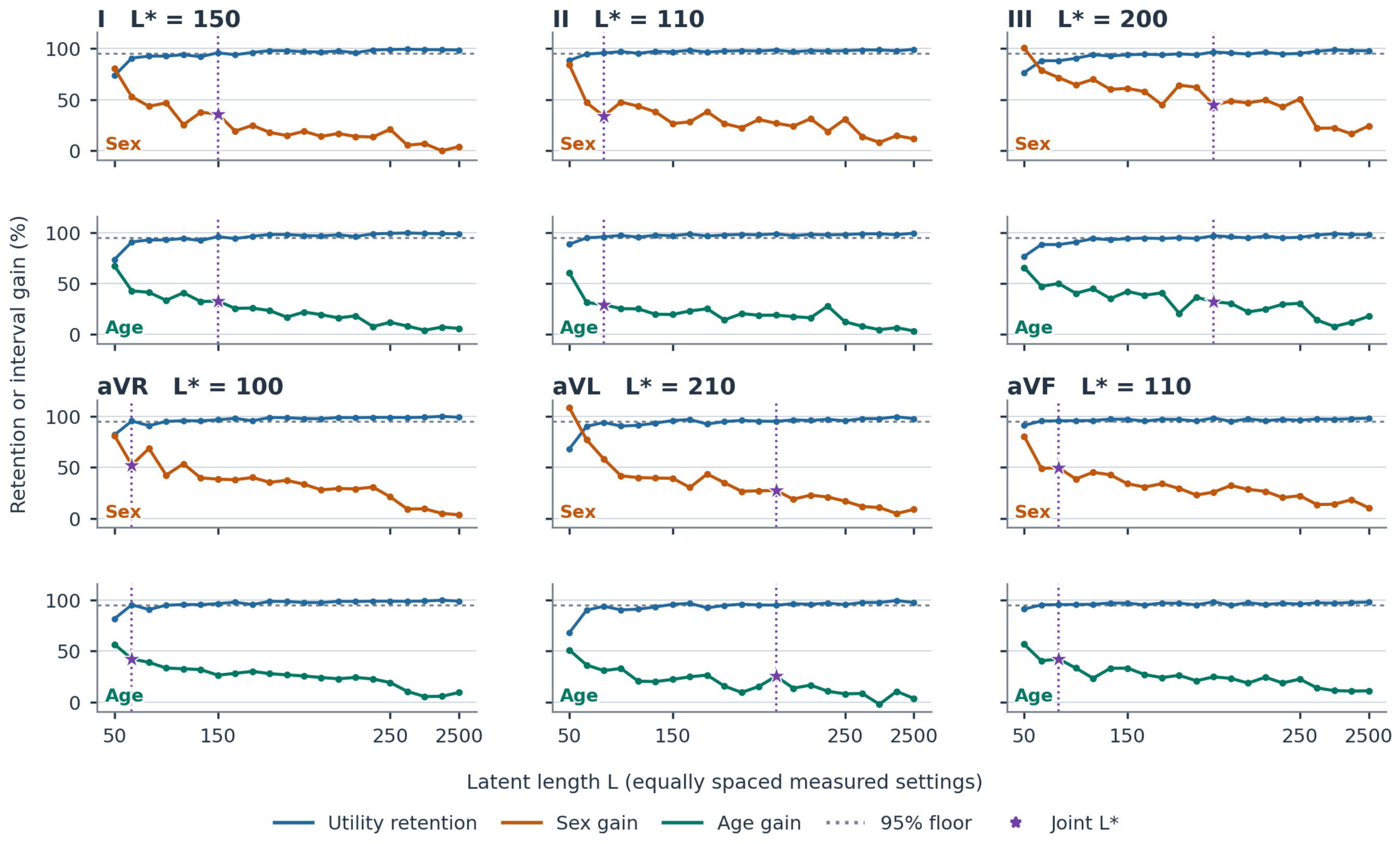

**Fig. 3.** Limb leads. Each lead has a sex panel and an age panel with the same utility curve. The star marks the privacy gain at the joint choice, and the vertical line locates it on both curves. The horizontal dotted line marks 95% utility retention. All 21 measured lengths appear in order with equal spacing. The dense central sequence is 100-250 in steps of ten.

## V. Operating points across the twelve leads

*A. A common tolerance selects different lengths*

Clean lead medians gave cross-lead medians of F = 0.674, S = 0.792 and E = 11.060 years. The nine clean scores alone still yielded S = 0.673 and E = 13.886 years. The clean-to-output gap ranged from 0.082 to 0.164 AUC units and from 2.178 to 3.572 years across leads. The same raw change can therefore yield different interval gains across leads.

For the full-length stochastic processor, the median lead retained 98.00% of clean utility and closed 8.72% of the sex and 10.33% of the age gap. At L = 250 the corresponding values were 96.80%, 21.05% and 20.68%. At L = 50 they were 73.87%, 80.49% and 60.27%. Supplementary Table S6 provides the equal-lead median diagnostic utility, demographic inference performance, privacy gains, and latent Kullback-Leibler (KL) divergence across all 21 tested temporal lengths. Table I reports the 12 joint choices. Their cross-lead raw medians were F = 0.648, S = 0.744 and E = 11.952 years. Separately computed medians of the lead-normalized U, $P_{sex}$ and $P_{age}$ were 95.74%, 34.98% and 32.46%, respectively.

*B. Limb leads*

In Fig. 3, lead I selected 150 for both attributes. In II, 110 gave a more even trade-off: moving to the sex-optimal 120 raised sex gain from 33.97% to 47.66% but lowered age gain from 29.27% to 25.38%. III selected 200; its sex-optimal 250 instead yielded 50.74% for sex and 30.60% for age. For aVR, 100 improved both targets substantially. Extending to its sex-optimal 130 changed the gains from 52.28/42.44% to 53.45/32.73%. For aVL, 210 retained 95.07% utility; selecting 150 for sex yielded 39.28/22.34%. Lead aVF selected 110 for both targets. Table S3 gives the measured utility and both gains at every joint or attribute-specific choice.

TABLE I
JOINT OPERATING POINTS WITH AT LEAST 95% OF CLEAN MACRO-F1 RETAINED

| Lead | L* | Macro-F1 ↑ | ΔU (pp) ↑ | Sex AUC* ↓ | Sex gain (%) ↑ | Age MAE (y) ↑ | Age gain (%) ↑ |
|---|---|---|---|---|---|---|---|
| **I** | 150 | 0.645 | −4.13 | 0.749 | 36.00 | 11.480 | 32.78 |
| **II** | 110 | 0.647 | −4.29 | 0.733 | 33.97 | 11.712 | 29.27 |
| **III** | 200 | 0.596 | −3.28 | 0.716 | 44.77 | 12.492 | 32.14 |
| **aVR** | 100 | 0.667 | −4.66 | 0.723 | 52.28 | 11.793 | 42.44 |
| **aVL** | 210 | 0.581 | −4.93 | 0.740 | 27.37 | 12.009 | 25.45 |
| **aVF** | 110 | 0.619 | −4.42 | 0.707 | 49.55 | 12.535 | 42.33 |
| **V1** | 110 | 0.628 | −4.24 | 0.738 | 47.59 | 12.654 | 47.62 |
| **V2** | 500 | 0.649 | −3.34 | 0.816 | 22.64 | 12.212 | 35.12 |
| **V3** | 240 | 0.653 | −4.83 | 0.826 | 14.16 | 11.895 | 22.97 |
| **V4** | 500 | 0.668 | −3.22 | 0.835 | 7.55 | 11.419 | 21.07 |
| **V5** | 140 | 0.678 | −3.66 | 0.793 | 24.56 | 11.645 | 29.91 |
| **V6** | 100 | 0.650 | −4.65 | 0.747 | 39.36 | 12.027 | 40.45 |

Raw values are fold medians. U = 100 + ΔU; ΔU is the change in normalized utility in percentage points (pp), not the raw macro-F1 difference. Sex and age gains each use that lead's own clean and score-only reference gap.

### *C. Precordial leads*

Fig. 4 separates two anterior leads with similar selected utility. V1 at 110 closed about 48% of both gaps. V2 required 500 for its joint choice and retained only 57.05% at 100. V3 selected 240, with a sex gain of 14.16% compared with an age gain of 22.97%. Both attributes selected 500 in V4; the sex gain was only 7.55%. V5 preferred 140 under the joint rule. Increasing L to its age-optimal 160 raised age gain from 29.91% to 33.32% and lowered sex gain from 24.56% to 23.83%. In V6, L = 100 closed approximately 40% of both gaps. In V2, V3 and V4, even the best eligible sex setting left AUC* above 0.81.

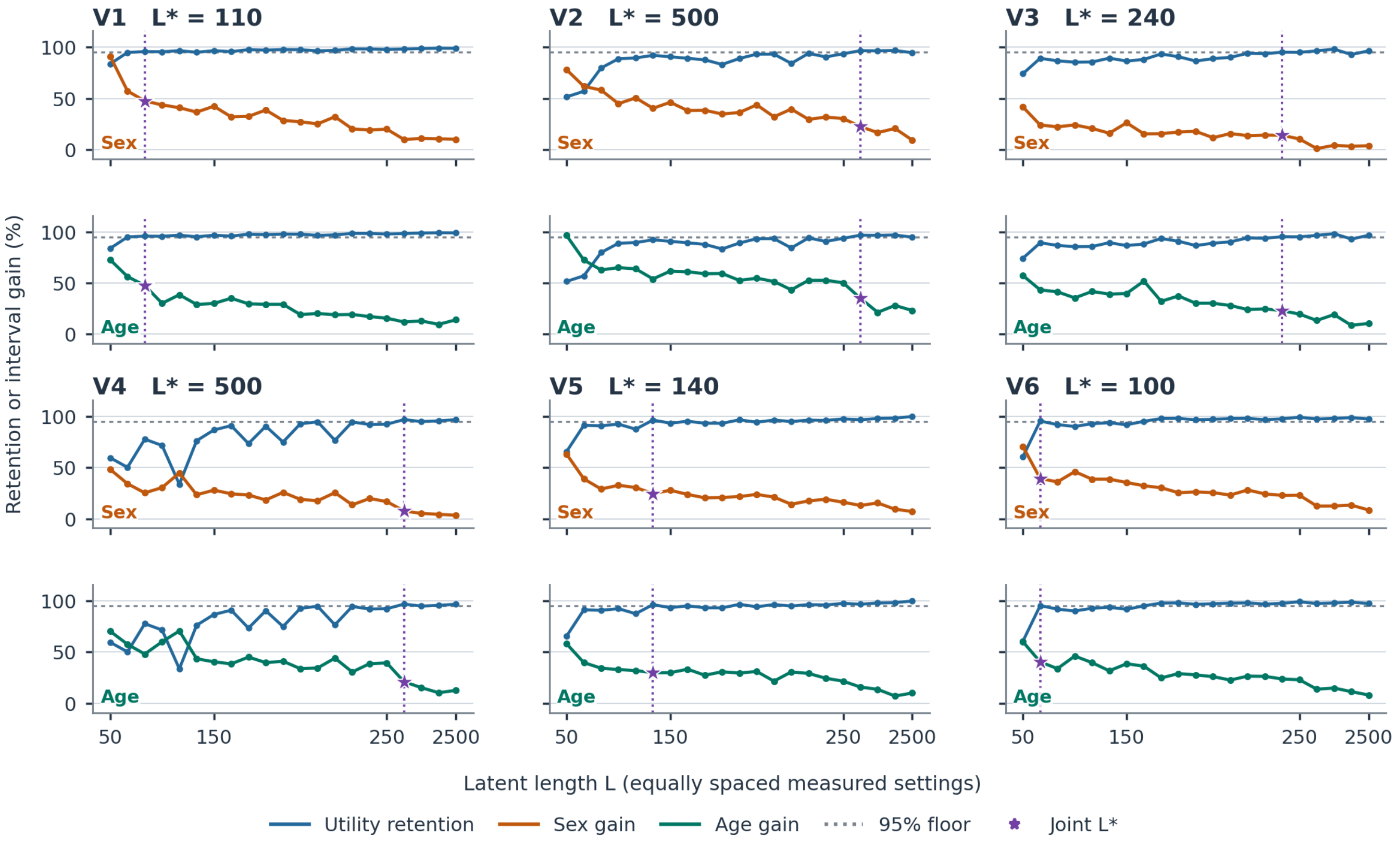

**Fig. 4.** Precordial leads with the same scales, order and symbols as Fig. 3. V1 and V6 retain high relative macro-F1 for the fixed HMC at short lengths. The joint choices for V2 and V4 occur at 500, with visibly different residual privacy gains.

## VI. What TIBECG adds over conventional transformations

### *A. Comparing privacy at the same utility floor*

A fixed minimum utility gives the device designer a concrete comparison. For each lead, we compared the joint TIBECG setting with the best of eight conventional transformations that reached U ≥ 95% on that lead. We repeated this separately for sex, age and their minimum J. Both methods must clear the same floor; their actual utility can differ. The eight settings and 21 lengths are discrete tested candidates. No interpolation or curve fitting enters the comparison.

For J, the joint TIBECG choice exceeded the best eligible conventional setting on all 12 leads. The median within-lead advantage was 22.08 percentage points (range 7.03-49.39). A Gaussian noise SD of 0.25 was the best conventional choice for I, II, III, aVR and aVL. Six-bit quantization was best for aVF and V1-V5. Laplace scale 0.15 was best for V6. Table S4 identifies, for each endpoint and lead, the winning conventional setting, its retained utility and its privacy score. This comparison conditions the claim on the stated utility requirement.

In Figs. 5 and 6, points nearer the upper right combine greater privacy improvement with higher utility. The frontier includes the clean waveform. The score-only reference is shown separately because it represents local inference rather than the assumed cloud-waveform deployment path. A joint choice can favor one attribute enough to lie below the frontier for the other. Supplementary Table S5 reports the utility and both privacy gains for every evaluated waveform condition on each lead, together with the sex- and age-specific Pareto-frontier membership underlying Figs. 5 and 6.

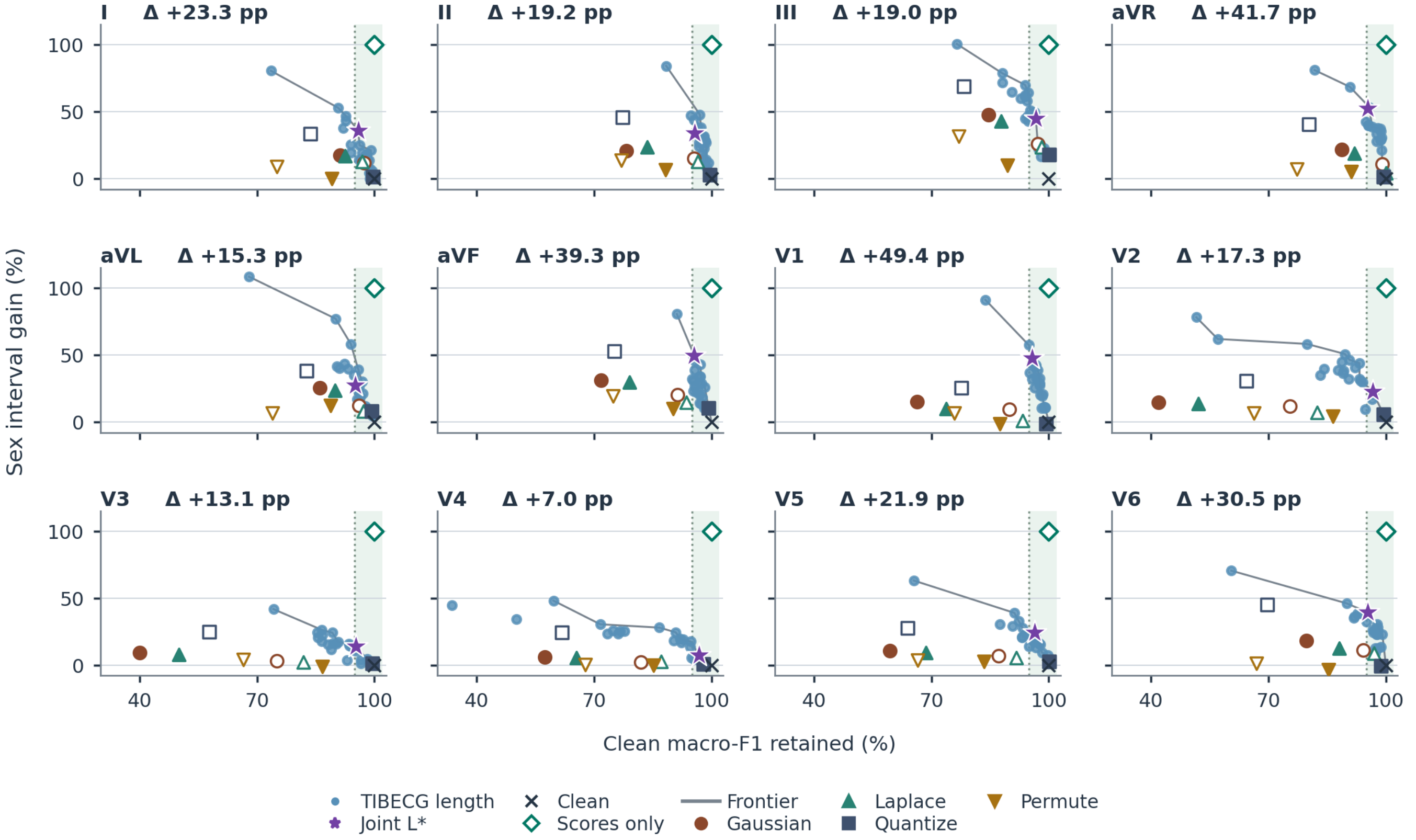

**Fig. 5.** Sex privacy versus utility. Each panel includes the 21 TIBECG lengths and all eight conventional settings. Open/filled markers respectively denote Gaussian SDs of 0.25/0.50 and Laplace scales of 0.15/0.30 times the record SD, quantization 3/6 bits and permutation 1/2 seconds. The Laplace noise SDs are approximately 0.212/0.424 times the record SD. The green region clears the 95% utility floor. Δ is the joint star's sex gain minus the best eligible conventional sex gain, in percentage points (pp). All waveform points enter the frontier computation.

### *B. What the sex comparison shows*

At the common floor the median advantage in sex gain was 20.55 points, ranging from 7.03 in V4 to 49.39 in V1. The selected point lay on the sex frontier in nine leads. In II, aVR and aVL another measured waveform offered a better sex-utility trade-off; the joint rule kept the length with the better weaker-target gain. For V4, six-bit quantization retained 98.00% of clean F1 but closed only 0.51% of the sex interval. TIBECG retained 96.78% and closed 7.55%. The absolute advantage on this lead was modest, even though the common-floor comparison favored TIBECG.

### C. What the age comparison shows

For age, the median advantage over the best eligible conventional value was 25.16 percentage points (range 10.62-48.78). All 12 joint selections lay on their age-utility frontier (Fig. 6). For lead III, the strongest eligible conventional age result came from Gaussian noise at 0.25 SD: 97.21% utility and 21.52% age gain. TIBECG retained 96.72% and achieved 32.14% age gain. This lead illustrates the smallest age advantage. At V1 the two approaches separated much more strongly even though both met the same threshold.

The conventional results illustrate how the transformation acts on the signal. Six-bit quantization leaves sample times unchanged and maps amplitudes to discrete levels through a nondecreasing mapping. Its diagnostic utility does not establish preservation of clinically relevant morphology. Across leads it retained a median 99.44% of clean utility but closed only 1.92% of the sex gap and 0.73% of the age gap. With three bits, gains rose to 35.63% and 41.62%, while retention fell to 76.27%. Independent noise can damage features needed by both readers. Raising Gaussian SD from 0.25 to 0.50 reduced median utility from 92.75% to 74.97%. Laplace scale 0.15 retained 95.02% for median gains of only 7.44% and 6.40%.

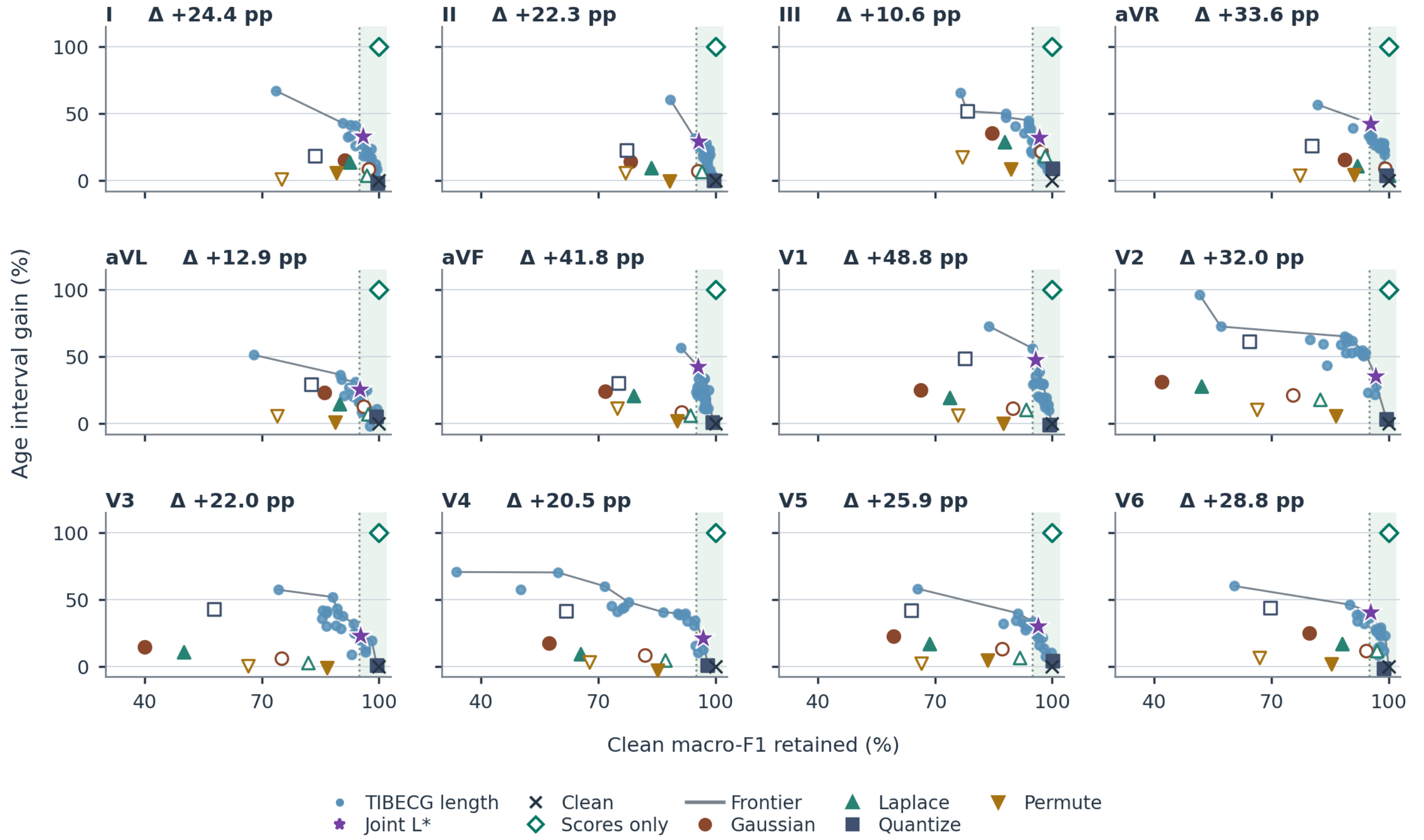

**Fig. 6.** Age privacy versus utility. Δ compares age gain at the joint star with the best eligible conventional age gain on the same lead. Gaussian SDs, Laplace scales, other marker settings, utility floor and frontier rule match Fig. 5. A higher age gain corresponds to a larger error in predicted age. Table S4 gives both coordinates of each common-floor comparison.

One-second permutation retained 74.35% utility while closing 6.52% of the sex gap and 5.38% of the age gap. Two-second blocks retained 87.86% with gains of 3.50% and 1.60%. Reordering blocks preserves shape inside a block and disrupts its wider temporal context. TIBECG extracts full-rate features before temporal pooling; protector weights are fitted jointly using diagnostic-score consistency. The observed frontiers are consistent with this design preserving features the teacher uses while making some demographic cues harder to read. They do not localize a cue or isolate the effect of an individual network component.

## VII. Interpreting the latent representation

### A. What the Gaussian diagnostic measures

The classical information bottleneck trades information about the input against information about a relevant target [29, author manuscript, p. 9]. Here "temporal information bottleneck" denotes the architectural restriction on L. We measure divergence after training. For d = 64L, posterior means $\mu_i$, log variances $v_i$ and a standard Gaussian reference r, the per-record divergence and its coordinate average are

$$K(x) = \tfrac{1}{2} \sum_i \left[\mu_i^2 + \exp(v_i) - 1 - v_i\right] \quad (4)$$

$$k(x) = K(x)/(64L).$$

The sum in (4) spans i = 1, …, d. We average K and k across test records within each fold, then take the median of ten fold means per lead and L. Equal-lead medians supply the overall curves in Fig. 7, with the corresponding numerical values for all 21 temporal lengths reported in Table S6. Supplementary Fig. S1 identifies every lead's trajectory relative to its full-length value. Natural logarithms give units of nats; one nat is 1/ln 2 bits of log-density ratio. An element with a shifted mean or a variance different from one contributes to K even if it carries no information about age or sex.

For a genuinely sampled latent Z with posterior q(z|x), I(X; Z) is the expected log ratio of q(z|x) to the aggregate posterior q(z). Adding and subtracting log q(z) yields $E_xK(x) = I(X; Z) + KL(q(z)\|r(z))$, the basis of the variational upper bound [30, p. 4, Eq. (14)]. The computation supplies only the left-hand quantity under that posterior model. It cannot separate input dependence from mismatch to r, nor recover attribute-specific mutual information. The keyed draw and the input-scale bypass in Fig. 1 also distinguish the released waveform from this ideal sampled latent.

*B. Separating fewer coordinates from larger per-coordinate divergence*

At fixed L, K = 64Lk, including after the fold medians. Consequently $K^*/K_{2500}$ is the product of L*/2,500 and $k^*/k_{2500}$ for a given lead. Shrinking the latent from 2,500 to 50 positions leaves 2% of the coordinates. Across leads median k rose from 3.170 to 4.305 nats/element. The 1.358-fold rise offsets part of the geometric reduction: median K falls to 2.72% of its full-length value (507,215 to 13,776 nats/record). At L = 250 the medians were 64,429 and 4.027, respectively. Thus fewer positions and stronger departures per surviving coordinate can occur together.

At L*, K fell by 74.92-94.93% and k rose by 14.38-42.90% versus each lead's full-length protector (Table II). Across the 21 settings, the median within-lead Spearman correlations of K with F, S and E were 0.860, 0.940 and −0.920.

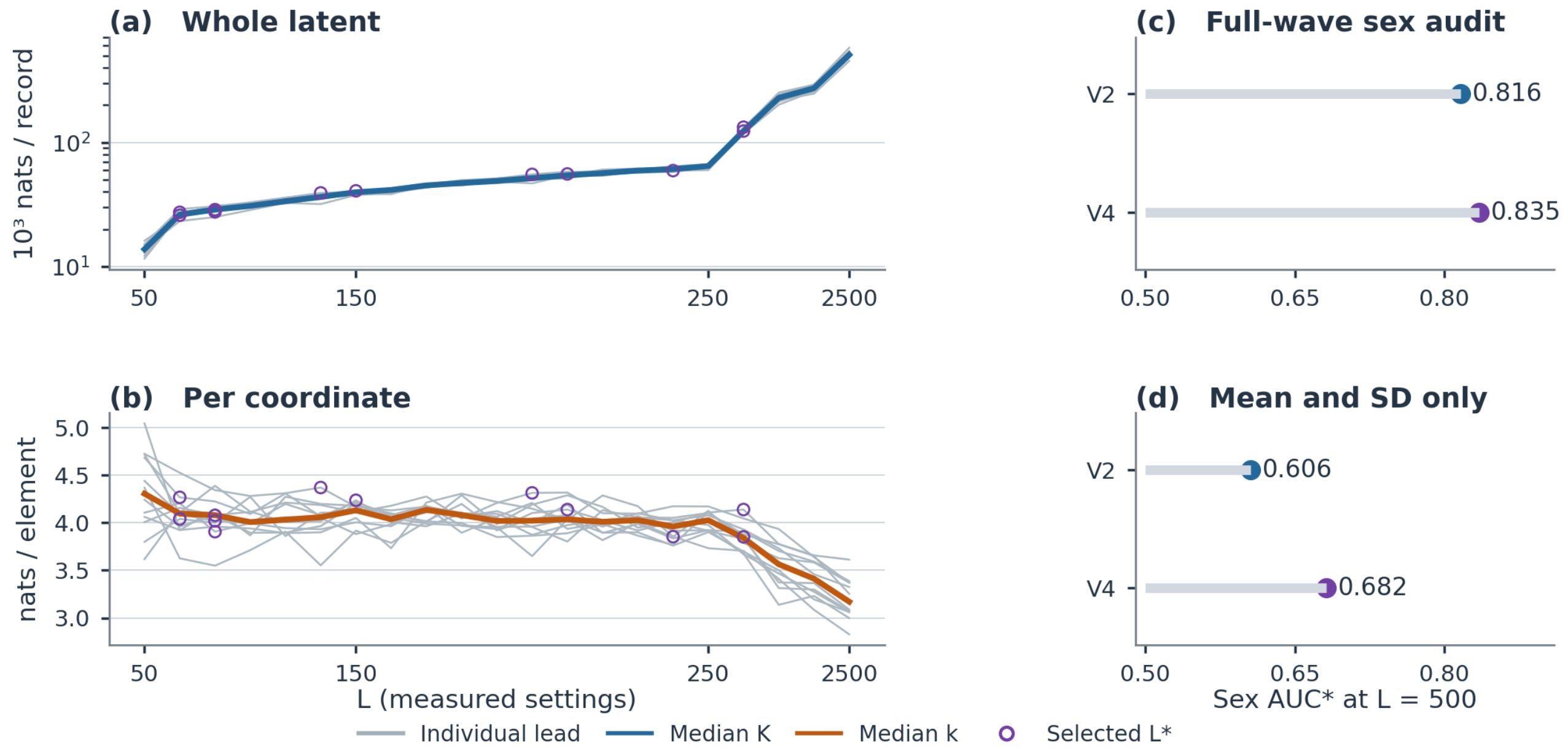

**Fig. 7.** Divergence and waveform evidence. (a-b) Twelve lead-specific trajectories, their equal-lead median and the selected points. K uses a log vertical scale; k uses a linear one. (c-d) At the same L and dimension, sex remains more predictable from V4 than V2 for both the full-wave and the two-feature audit. The two-feature model uses the released waveform's mean and population SD.

TABLE II
LATENT DIVERGENCE AND TWO-FEATURE SEX AUDIT AT THE SAME OPERATING POINTS

| Lead | L* | K (10³ nats/record) | k (nats/element) | K drop (%) | k change (%) | Mean/SD AUC* |
|---|---|---|---|---|---|---|
| **I** | 150 | 40.67 | 4.236 | 91.70 | +38.31 | 0.548 |
| **II** | 110 | 28.31 | 4.021 | 94.27 | +30.25 | 0.578 |
| **III** | 200 | 55.22 | 4.314 | 89.76 | +27.97 | 0.552 |
| **aVR** | 100 | 27.31 | 4.267 | 94.31 | +42.36 | 0.572 |
| **aVL** | 210 | 55.62 | 4.139 | 89.31 | +27.21 | 0.532 |
| **aVF** | 110 | 27.50 | 3.906 | 94.93 | +15.33 | 0.548 |
| **V1** | 110 | 28.71 | 4.078 | 94.60 | +22.67 | 0.562 |
| **V2** | 500 | 132.44 | 4.139 | 77.09 | +14.57 | 0.606 |
| **V3** | 240 | 59.19 | 3.854 | 89.02 | +14.38 | 0.689 |
| **V4** | 500 | 123.25 | 3.852 | 74.92 | +25.42 | 0.682 |
| **V5** | 140 | 39.16 | 4.370 | 92.00 | +42.84 | 0.656 |
| **V6** | 100 | 25.86 | 4.040 | 94.28 | +42.90 | 0.583 |

K drop = 100(1 − K*/K at 2,500). k change = 100(k*/k at 2,500 − 1). Positive values give the magnitude of the decrease in K and increase in k. The final column audits two release statistics.

For k they were −0.455, −0.525 and 0.538 (Table S7). All three outcomes can co-vary with L. These descriptive correlations alone do not identify which latent coordinates carry demographic information.

*C. A fixed-length test of the interpretation*

V2 and V4 selected L = 500, or 32,000 latent elements. V4 had lower K (123.25 versus 132.44 thousand nats) and lower k (3.852 versus 4.139), yet its released waveform had higher sex AUC* (0.835 versus 0.816). Its sex gap closure was smaller (7.55% versus 22.64%). Holding the number of coordinates constant exposes a ranking reversal. Lower divergence to this prior is insufficient to rank the two leads by residual sex leakage.

We checked the same comparison with a separately fitted recipient that uses only the released waveform's mean and population SD. The two features were standardized on the training split. Logistic regression predicted sex and ridge regression predicted age. At L = 500 the two-feature sex AUC* was 0.682 for V4 and 0.606 for V2. The corresponding age MAEs were 15.677 and 15.671 years, compared with full-wave values of 11.419 and 12.212. The released summary therefore contains a measurable sex cue under this attack. It is derived from the decoder output and the rescaling path together. It cannot determine what fraction, if any, came from the bypass alone. A bypass ablation or localized waveform analysis would answer a different question.

## VIII. Acquisition rate and selection stability

*A. Utility and privacy at matched time counts*

Fig. 8 compares the trade-off from reducing internal time positions with that from reducing the observed waveform samples. Independent HMCs and three-seed age/sex auditors were fitted at each measured rate (f); the 250-Hz row reuses the clean reference. We paired seven observed rates with L = 10f. TIBECG still reads and releases 2,500 samples, with 64 features per latent position. Each endpoint is normalized using that lead's clean and output-only references, as in (1)-(3). We subtract rate-baseline percentages from TIBECG percentages within each lead before taking cross-lead medians.

At 5, 10, 25 and 50 Hz, TIBECG retained more utility on all 12 leads, with median paired advantages of 43.93, 45.70, 14.53 and 5.27 percentage points. The corresponding sex-gain differences were −66.66, −39.21, −18.27 and −10.96 points, and age-gain differences were −46.33, −31.00, −18.97 and −6.73. Thus direct downsampling generally produced stronger demographic suppression, but at a substantially larger loss of diagnostic utility. TIBECG favored utility preservation within this privacy-utility trade-off rather than uniformly maximizing privacy. At 5 and 10 Hz, every lead showed this pattern. TIBECG therefore does not uniformly dominate downsampling, but generally preserves substantially more utility while accepting smaller demographic-suppression gains.

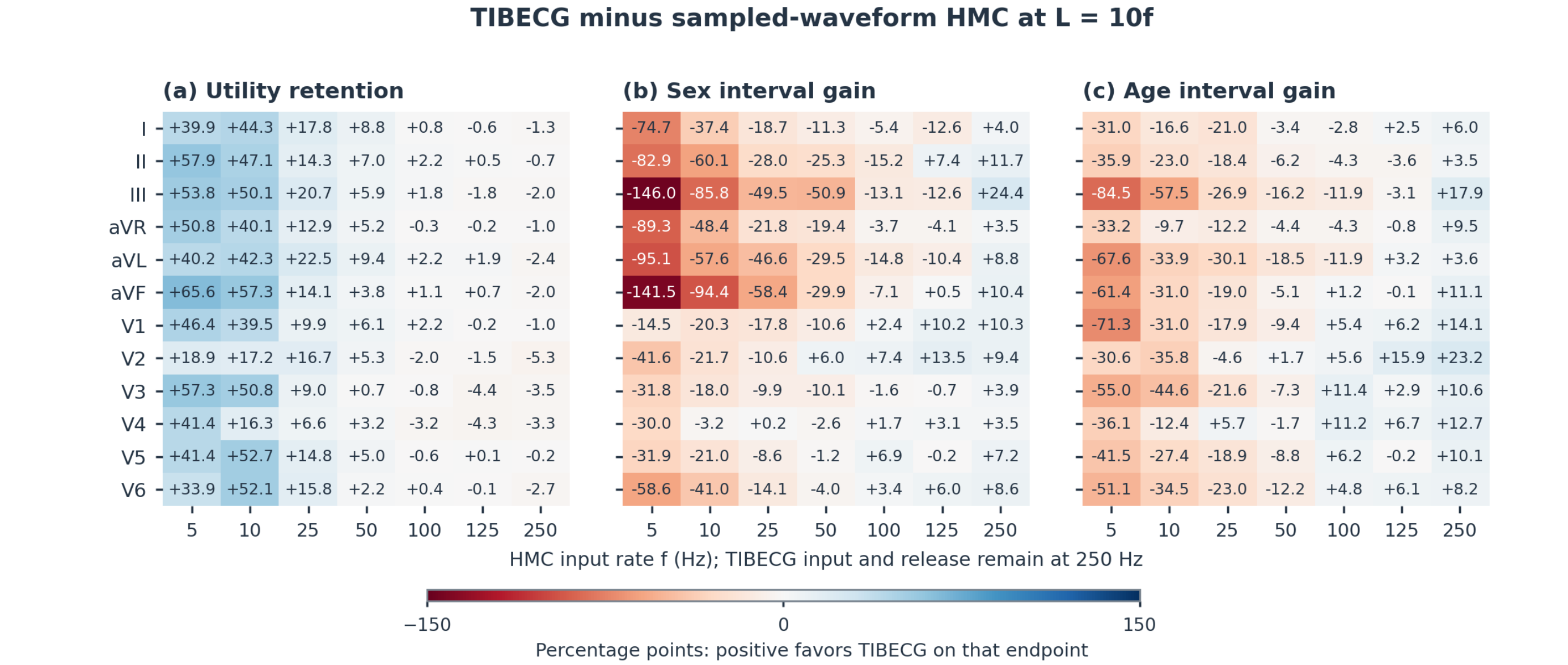

Fig. 8. Paired differences in utility retention, sex interval gain and age interval gain at L = 10f. Positive values favor TIBECG on that endpoint; negative values favor the sampled waveform. Cells are rounded to 0.1 percentage point; Supplementary Fig. S2 shows both absolute percentage curves per lead. Tables S9 and S10 provide paired values and equal-lead summaries.

The full-rate trunk can combine local morphology before temporal pooling; the rate baseline observes an already decimated waveform. This helps explain why TIBECG can reduce demographic inference without paying the full utility cost of direct rate reduction. The observed utility advantage is conditional on access to richer input. The fixed 41-tap filter designed for 500-to-250-Hz resampling was reused at every lower rate, so inadequate anti-alias filtering may disadvantage those baselines. A properly rate-matched filter is needed to separate sampling loss from this implementation effect. Matching L to 10f equates neither bandwidth, representation size nor transmission cost. Table S10 also shows sex gain above 100% at 5 Hz: the tested attacker performs worse than the score-only reference, not beyond a universal privacy bound. Energy-aware ECG classification instead optimizes accuracy and energy jointly [31, p. 4739, Eq. (8)]. We recorded transform timings in the experimental compute environment, but did not evaluate on-device latency, energy consumption or hardware feasibility.

### B. Lead-specific sensitivity of the selected length

Fig. 9 tests tolerance dependence and sensitivity to the available test folds. We repeat the maximin rule at 90%, 95% and 97.5% utility floors, then omit each fold and recompute both references, every median and the 95% choice. Table S1 gives all selections and near-optimal alternatives at the 95% floor: other feasible lengths with J no more than five percentage points below the selected value. Table S11 distinguishes an original choice becoming ineligible from another length overtaking it; it also reports the original utility margin and the gap to the next-best eligible J.

V6 reproduced L = 100 in 10/10 omissions; aVF and V5 reproduced their choices in 9/10, and III in 8/10. aVL and V2 reproduced theirs in only 2/10. These cases have different explanations. aVL is only 0.07 percentage point above the utility floor and 0.59 point ahead in J; its original choice remains feasible in 5/10 omissions. V2 remains feasible in all ten, but competing lengths overtake it, with selections ranging from 230 to 1,250. V3 and V4 range up to 2,500. At 97.5%, neither V2 nor V4 has an eligible candidate. Thus some choices are more stable under this particular perturbation; the result does not establish intrinsic lead robustness or transfer to noisy devices. Shared training records [32, p. 1089] and post hoc selection preclude interpreting these counts as independent replication.

| | Utility floor (%) | | | Omitted fold (95% utility floor) | | | | | | | | | | Same L* |
|---|---|---|---|---|---|---|---|---|---|---|---|---|---|---|
| | 90 | 95 | 97.5 | 0 | 1 | 2 | 3 | 4 | 5 | 6 | 7 | 8 | 9 | n / 10 |
| I | 100 | **150** | 180 | 150 | 250 | 170 | 170 | 150 | 170 | 150 | 150 | 180 | 200 | **4** |
| II | 100 | **110** | 160 | 110 | 110 | 130 | 110 | 100 | 100 | 100 | 100 | 100 | 110 | **4** |
| III | 130 | **200** | 2500 | 200 | 130 | 200 | 200 | 200 | 160 | 200 | 200 | 200 | 200 | **8** |
| aVR | 100 | **100** | 160 | 130 | 100 | 100 | 100 | 180 | 100 | 100 | 100 | 120 | 100 | **7** |
| aVL | 100 | **210** | 500 | 160 | 160 | 160 | 160 | 160 | 210 | 150 | 160 | 210 | 160 | **2** |
| aVF | 50 | **110** | 200 | 110 | 110 | 110 | 110 | 110 | 100 | 110 | 110 | 110 | 110 | **9** |
| V1 | 100 | **110** | 170 | 110 | 110 | 110 | 100 | 100 | 110 | 110 | 100 | 100 | 100 | **5** |
| V2 | 150 | **500** | — | 500 | 1250 | 230 | 230 | 1250 | 250 | 250 | 500 | 1250 | 230 | **2** |
| V3 | 180 | **240** | 1000 | 2500 | 1000 | 2500 | 250 | 230 | 240 | 240 | 240 | 240 | 1000 | **4** |
| V4 | 160 | **500** | — | 210 | 210 | 230 | 2500 | 500 | 500 | 200 | 500 | 500 | 210 | **4** |
| V5 | 100 | **140** | 250 | 140 | 200 | 140 | 140 | 140 | 140 | 140 | 140 | 140 | 140 | **9** |
| V6 | 120 | **100** | 220 | 100 | 100 | 100 | 100 | 100 | 100 | 100 | 100 | 100 | 100 | **10** |

Fig. 9. Utility-tolerance and fold-omission sensitivity of the selected length. A dash means no candidate meets that floor. Shaded omission cells repeat the full-data 95% choice; the last column counts agreement. Table S1 contains all selections and near-optimal alternatives, and Table S11 explains feasibility and ranking changes. No models were retrained for this analysis.

## IX. Physiological interpretation and scope

### A. Sex information in anterior leads

On clean ECG the sex AUC* was 0.847 in V2, 0.849 in V3 and 0.847 in V4. The excesses over the median of the other 11 leads were 0.056, 0.058 and 0.056. At L = 250 they were 0.040, 0.066 and 0.054. At L = 50, absolute AUC* fell to 0.741, 0.781 and 0.768 but the excesses were 0.049, 0.088 and 0.076. The relative ranking can persist, or even widen, while the sex auditor becomes less discriminative in absolute AUC*. Table S8 gives the actual scores and comparison medians at every measured length.

Sex and age differences in ECG amplitudes and intervals have been measured in adults [33]. ST amplitude in particular varies by age, sex and lead [34]. Endres et al. measured J-point and subsequent ST heights in the anterior group V2-V4 and found higher resting levels in men in their sample [35, Fig. 3]. V2 and V3 have sex- and age-dependent reference thresholds for ST elevation [36]. V4 samples the left anterior chest more apically [37, ch. 15]. These observations make repolarization and amplitude plausible contributors to the anterior signal. The pooled clinical measurements do not establish a separate ordering for V2, V3 and V4. Our two-feature audit tests gross waveform statistics; it does not identify the ST segment as the source.

### B. Age information in aVR and I

On clean ECG, aVR and I yielded age MAEs of 10.277 and 10.455 years, lower than the other-lead medians by 0.938 and 0.761 years, respectively. At L = 250 the advantages were 0.834 and 0.969 years. At L = 50 the errors rose to 12.303 and 12.555 years and the advantages narrowed to 0.599 and 0.347. The joint settings left aVR at 11.793 years and I at 11.480. Lead I measures a leftward frontal projection and aVR faces the right shoulder [37, ch. 15]. The Framingham reference sample showed a leftward shift of the QRS axis with age [38]. A change in the direction of the ventricular activation vector could alter these two projections. Diagnosis mix and electrode position could also influence the errors. The present audits did not estimate an individual's QRS axis or test waveform-region masking.

### C. Transfer hypothesis and retrospective selection

Two prior studies involving the present authors used the same five-block CNN-bidirectional GRU-attention backbone for voice [39] and cough [40, Table 1, note c], extending it with demographic or clinical inputs for multimodal prediction. This backbone matches the current HMC architecture and motivates fitting an analogous processor to an acoustic teacher and auditing speaker attributes. Acoustic utility, privacy and microphone noise require separate evaluation.

The ECG choices are retrospective: selecting and reporting L on the same test summaries can favor optimistic settings [41, pp. 2079-2080]. Validation selected checkpoints only. Unequal candidate counts (21 lengths versus eight transformations) also limit comparator fairness.

## X. Limitations

*A. Clinical acquisition and consumer use*

All recordings came from hospital twelve-lead acquisition, although each model used only one extracted lead. Many consumer devices record a single lead in ambulatory conditions, where motion and electrode contact can introduce artifacts [42, Sec. 2.2.1]. Processing hospital channels separately does not reproduce consumer electrodes, filtering or noise. Clinical-to-home single-lead models can lose performance across devices [43]. We tested neither real consumer recordings nor controlled motion, baseline drift, contact noise or placement changes. V2-V4 findings cannot be transferred to a consumer lead-I recording. Utility and leakage require device-specific, patient-separated external evaluation; the direction and magnitude of either change remain unknown.

*B. Waveform fidelity and expert interpretation*

No cardiologist or other human expert evaluated the transformed waveforms. Score consistency does not constrain reconstruction error, ECG intervals or morphology; different traces can yield similar scores. Thus 95% macro-F1 retention is neither 95% accuracy nor evidence of cardiologist-readable reconstruction. Expert-annotated ECG studies evaluate wave representation, fiducial timing and intervals separately [44]. Before proposing human review, blinded readers should assess paired original and transformed traces for rhythm and P-QRS-T morphology, PR/QRS/QT and ST measurements, unreadable-trace rates and inter-reader agreement. These evaluations remain future work.

*C. When waveform transmission is required*

If the intended classifier can run locally, transmitting its scores or class label suffices and TIBECG is unnecessary. Sending predicted classes has also been implemented in an ECG monitoring prototype [8, p. 111]. Our motivating deployment setting is different: the waveform-input classifier remains a proprietary cloud-service asset and is not deployed on the consumer device. The waveform must therefore be transmitted to obtain the service's diagnostic scores.

The output-only reference represents the lower-exposure case that would be possible after local inference; it is not the assumed deployment path. Output-only scores yield exactly the clean class decisions and macro-F1 by construction, yet still leak demographics (equal-lead median sex AUC* 0.673; age MAE 13.886 years). Class-label-only leakage was not audited. Clinician overread [42, Sec. 2.2.1], morphology measurements or different downstream models can provide additional reasons for waveform access, but none was validated on these transformed traces. TIBECG therefore addresses the conditional case in which a waveform must reach a remote classifier while reducing information unnecessary for its fixed diagnostic task.

## XI. Conclusion

TIBECG reduces two measured demographic attacks while retaining much of a fixed classifier's utility on individual clinical leads. At a 95% utility floor, all 12 joint choices outperformed eligible conventional transformations on the weaker privacy gain. The implemented sampling-rate comparison shows a different trade-off: direct downsampling generally achieved stronger demographic suppression but at a substantially larger loss of diagnostic utility, whereas TIBECG favored utility preservation. Fold omissions expose selection sensitivity. Latent KL does not replace release-level auditing. These empirical audits do not constitute a formal privacy guarantee; stronger attacks or repeated releases may reveal additional leakage. Consumer-device and expert waveform validation remain necessary. When inference can run locally, score-only transmission remains preferable; TIBECG instead targets the conditional setting in which a proprietary cloud classifier requires waveform input.

## Acknowledgment

The authors would like to thank the Academia Sinica Artificial Intelligence Cooperative (AS-IAIA-114-AI01) for conceptual support. OpenAI ChatGPT (Codex) assisted with English editing, code implementation for statistical analyses, and preparation of figures and tables. The authors independently verified the analyses and retain responsibility for the final content.

## References

[1] G. Biagetti, P. Crippa, L. Falaschetti, S. Orcioni, and C. Turchetti, "Wireless surface electromyograph and electrocardiograph system on 802.15.4," IEEE Trans. Consum. Electron., vol. 62, no. 3, pp. 258-266, 2016. doi 10.1109/TCE.2016.7613192

[2] N. Dey, A. S. Ashour, F. Shi, S. J. Fong, and R. S. Sherratt, "Developing residential wireless sensor networks for ECG healthcare monitoring," IEEE Trans. Consum. Electron., vol. 63, no. 4, pp. 442-449, 2017. doi 10.1109/TCE.2017.015063

[3] W.-C. Kao, W.-H. Chen, C.-K. Yu, C.-M. Hong, and S.-Y. Lin, "Portable real-time homecare system design with digital camera platform," IEEE Trans. Consum. Electron., vol. 51, no. 4, pp. 1035-1040, 2005. doi 10.1109/TCE.2005.1561822

[4] G. Tangari, M. Ikram, K. Ijaz, M. A. Kaafar, and S. Berkovsky, "Mobile health and privacy: cross sectional study," BMJ, vol. 373, Art. n1248, 2021. doi 10.1136/bmj.n1248

[5] Z. I. Attia et al., "Age and Sex Estimation Using Artificial Intelligence From Standard 12-Lead ECGs," Circ. Arrhythm. Electrophysiol., vol. 12, no. 9, Art. e007284, 2019. doi 10.1161/CIRCEP.119.007284

[6] T.-M. Chen et al., "SRECG: ECG Signal Super-Resolution Framework for Portable/Wearable Devices in Cardiac Arrhythmias Classification," IEEE Trans. Consum. Electron., vol. 69, no. 3, pp. 250-260, 2023. doi 10.1109/TCE.2023.3237715

[7] S. Raj and K. C. Ray, "A Personalized Point-of-Care Platform for Real-Time ECG Monitoring," IEEE Trans. Consum. Electron., vol. 64, no. 4, pp. 452-460, 2018. doi 10.1109/TCE.2018.2877481

[8] S. Raj, "An Efficient IoT-Based Platform for Remote Real-Time Cardiac Activity Monitoring," IEEE Trans. Consum. Electron., vol. 66, no. 2, pp. 106-114, 2020. doi 10.1109/TCE.2020.2981511

[9] S.-Y. Lee, P.-W. Huang, M.-C. Liang, J.-H. Hong, and J.-Y. Chen, "Development of an Arrhythmia Monitoring System and Human Study," IEEE Trans. Consum. Electron., vol. 64, no. 4, pp. 442-451, 2018. doi 10.1109/TCE.2018.2875799

[10] P. Maji, H. K. Mondal, A. P. Roy, S. Poddar, and S. P. Mohanty, "iKardo: An Intelligent ECG Device for Automatic Critical Beat Identification for Smart Healthcare," IEEE Trans. Consum. Electron., vol. 67, no. 4, pp. 235-243, 2021. doi 10.1109/TCE.2021.3129316

[11] N. Sharma and D. Joshi, "DRGCN-BiLSTM: An Electrocardiogram Heartbeat Classification Using Dynamic Spatial-Temporal Graph Convolutional and Bidirectional Long-Short Term Memory Technique," IEEE Trans. Consum. Electron., vol. 71, no. 1, pp. 579-593, 2025. doi 10.1109/TCE.2025.3540875

[12] A. Bilgin, M. W. Marcellin, and M. I. Altbach, "Compression of electrocardiogram signals using JPEG2000," IEEE Trans. Consum. Electron., vol. 49, no. 4, pp. 833-840, 2003. doi 10.1109/TCE.2003.1261162

[13] S.-Y. Lee, P.-H. Su, Y.-W. Hung, I.-P. Lee, S.-J. Li, and J.-Y. Chen, "Motion Artifact Reduction Algorithm for Wearable Electrocardiogram Monitoring Systems," IEEE Trans. Consum. Electron., vol. 69, no. 3, pp. 533-547, 2023. doi 10.1109/TCE.2023.3279258

[14] M. A. Sayeed, S. P. Mohanty, E. Kougianos, and H. P. Zaveri, "ESeiz: An Edge-Device for Accurate Seizure Detection for Smart Healthcare," IEEE Trans. Consum. Electron., vol. 65, no. 3, pp. 379-387, 2019. doi 10.1109/TCE.2019.2920068

[15] I. L. Olokodana, S. P. Mohanty, E. Kougianos, and R. S. Sherratt, "EZcap: A Novel Wearable for Real-Time Automated Seizure Detection From EEG Signals," IEEE Trans. Consum. Electron., vol. 67, no. 2, pp. 166-175, 2021. doi 10.1109/TCE.2021.3079399

[16] L. Rachakonda, S. P. Mohanty, E. Kougianos, and P. Sundaravadivel, "Stress-Lysis: A DNN-Integrated Edge Device for Stress Level Detection in the IoMT," IEEE Trans. Consum. Electron., vol. 65, no. 4, pp. 474-483, 2019. doi 10.1109/TCE.2019.2940472

[17] L. Rachakonda, A. K. Bapatla, S. P. Mohanty, and E. Kougianos, "SaYoPillow: Blockchain-Integrated Privacy-Assured IoMT Framework for Stress Management Considering Sleeping Habits," IEEE Trans. Consum. Electron., vol. 67, no. 1, pp. 20-29, 2021. doi 10.1109/TCE.2020.3043683

[18] V. P. Yanambaka, S. P. Mohanty, E. Kougianos, and D. Puthal, "PMsec: Physical Unclonable Function-Based Robust and Lightweight Authentication in the Internet of Medical Things," IEEE Trans. Consum. Electron., vol. 65, no. 3, pp. 388-397, 2019. doi 10.1109/TCE.2019.2926192

[19] S. Prajapat, D. Gautam, P. Kumar, A. K. Das, and M. S. Hossain, "Designing Lattice-Based Sequential Aggregate Signature Scheme for Securing Consumer Electronics-Centric IoMT," IEEE Trans. Consum. Electron., vol. 71, no. 4, pp. 11676-11685, 2025. doi 10.1109/TCE.2025.3595948

[20] K. Park, S. Prajapat, and Y. Park, "Full Decentralized Privacy-Preserving Scheme for Data Access Control in Smart Consumer Electronic Devices for Medical IoT," IEEE Trans. Consum. Electron., vol. 72, no. 2, pp. 3237-3250, 2026. doi 10.1109/TCE.2026.3665715

[21] U. Islam, H. Ullah, N. Khan, I. Ahmad, and K. Saleem, "Adaptive Federated Learning Framework for Privacy-Preserving Consumer-Centric IoMT: A Novel Secure Data Collaboration Model," IEEE Trans. Consum. Electron., vol. 71, no. 4, pp. 10134-10151, 2025. doi 10.1109/TCE.2025.3606642

[22] J. Lv, B.-G. Kim, and K. Li, "Fed-HealthGen: A Generative Federated Framework for Privacy-Preserving Personalized Healthcare Using Wearable Consumer Electronics," IEEE Trans. Consum. Electron., vol. 71, no. 4, pp. 11440-11452, 2025. doi 10.1109/TCE.2025.3624693

[23] J. Wang, B.-G. Kim, S. Rani, K. Li, and J. Lv, "HealthGuard: Privacy-Preserving Framework for Consumer Healthcare Devices Against Adversarial Inference Attacks," IEEE Trans. Consum. Electron., vol. 72, no. 1, pp. 1465-1476, 2026. doi 10.1109/TCE.2025.3649226

[24] F. S. Shishir, C. J. Harvey, A. Gupta, A. Noheria, and S. Shomaji, "AI-enabled privacy-preserving cardiac diagnostics via electrocardiograms," Sci. Rep., vol. 16, Art. 17571, 2026. doi 10.1038/s41598-026-47665-6

[25] H. Edwards and A. Storkey, "Censoring representations with an adversary," Proc. Int. Conf. Learn. Represent., 2016. Public text

[26] F. Liu et al., "An Open Access Database for Evaluating the Algorithms of Electrocardiogram Rhythm and Morphology Abnormality Detection," J. Med. Imaging Health Inform., vol. 8, no. 7, pp. 1368-1373, 2018. doi 10.1166/jmihi.2018.2442

[27] T.-M. Chen, C.-H. Huang, E. S. C. Shih, Y.-F. Hu, and M.-J. Hwang, "Detection and Classification of Cardiac Arrhythmias by a Challenge-Best Deep Learning Neural Network Model," iScience, vol. 23, no. 3, Art. 100886, 2020. doi 10.1016/j.isci.2020.100886

[28] D. P. Kingma and M. Welling, "Auto-Encoding Variational Bayes," Proc. Int. Conf. Learn. Represent., 2014. Author text (arXiv v11)

[29] N. Tishby, F. C. Pereira, and W. Bialek, "The information bottleneck method," Proc. 37th Allerton Conf. Commun., Control, Comput., pp. 368-377, 1999. Public text

[30] A. A. Alemi, I. Fischer, J. V. Dillon, and K. Murphy, "Deep Variational Information Bottleneck," Proc. Int. Conf. Learn. Represent., 2017. Public text

[31] C. Liu, J. Liu, and D. Wang, "Energy-Efficient Health Monitoring in Smart Devices: Deep Learning-Based Technique for Wearable ECG Systems," IEEE Trans. Consum. Electron., vol. 71, no. 2, pp. 4736-4743, 2025. doi 10.1109/TCE.2025.3569456

[32] Y. Bengio and Y. Grandvalet, "No Unbiased Estimator of the Variance of K-Fold Cross-Validation," J. Mach. Learn. Res., vol. 5, pp. 1089-1105, 2004. Public text

[33] P. R. Rijnbeek et al., "Normal values of the electrocardiogram for ages 16-90 years," J. Electrocardiol., vol. 47, no. 6, pp. 914-921, 2014. doi 10.1016/j.jelectrocard.2014.07.022

[34] P. W. Macfarlane, "Age, sex, and the ST amplitude in health and disease," J. Electrocardiol., vol. 34, suppl., pp. 235-241, 2001. doi 10.1054/jelc.2001.28906

[35] S. Endres, K. A. Mayuga, A. de Cristofaro, T. Taneja, J. J. Goldberger, and A. H. Kadish, "Age and Gender Difference in ST Height at Rest and after Double Autonomic Blockade in Normal Adults," Ann. Noninvasive Electrocardiol., vol. 11, no. 3, pp. 253-258, 2006. doi 10.1111/j.1542-474X.2006.00112.x

[36] P. M. Rautaharju, B. Surawicz, and L. S. Gettes, "AHA/ACCF/HRS recommendations for the standardization and interpretation of the electrocardiogram: Part IV: the ST segment, T and U waves, and the QT interval," J. Am. Coll. Cardiol., vol. 53, no. 11, pp. 982-991, 2009. doi 10.1016/j.jacc.2008.12.014

[37] J. Malmivuo and R. Plonsey, Bioelectromagnetism: Principles and Applications of Bioelectric and Biomagnetic Fields. New York, NY, USA: Oxford Univ. Press, 1995, ch. 15. Public text

[38] D. Levy et al., "Electrocardiographic changes with advancing age. A cross-sectional study of the association of age with QRS axis, duration and voltage," J. Electrocardiol., vol. 20, suppl., pp. 44-47, 1987. Public record

[39] C.-T. Wang, T.-M. Chen, N.-T. Lee, and S.-H. Fang, "AI Detection of Glottic Neoplasm Using Voice Signals, Demographics, and Structured Medical Records," Laryngoscope, vol. 134, no. 11, pp. 4585-4592, 2024. doi 10.1002/lary.31563

[40] D. Jaganath et al., "Accelerating Cough-Based Algorithms for Pulmonary Tuberculosis Screening: Results From the CODA TB DREAM Challenge," Open Forum Infect. Dis., vol. 12, no. 10, Art. ofaf572, 2025. doi 10.1093/ofid/ofaf572

[41] G. C. Cawley and N. L. C. Talbot, "On Over-fitting in Model Selection and Subsequent Selection Bias in Performance Evaluation," J. Mach. Learn. Res., vol. 11, pp. 2079-2107, 2010. Public text

[42] N. Varma et al., "2021 ISHNE/HRS/EHRA/APHRS collaborative statement on mHealth in Arrhythmia Management: Digital Medical Tools for Heart Rhythm Professionals," J. Arrhythmia, vol. 37, no. 2, pp. 271-319, 2021. doi 10.1002/joa3.12461

[43] C. Lin et al., "Bridging the gap from clinical to home ECG: quantifying and overcoming accuracy loss in AI-enabled single-lead ECG models," npj Digit. Med., vol. 9, Art. 510, 2026. doi 10.1038/s41746-026-02919-5

[44] P. L. Rajbhandary et al., "ECG Signal Quality Assessments of a Small Bipolar Single-Lead Wearable Patch Sensor," Cardiovasc. Eng. Technol., vol. 13, no. 5, pp. 783-796, 2022. doi 10.1007/s13239-022-00617-3